# Orientation to Pose: Continuum Robots Shape Sensing Based on Piecewise Polynomial Curvature Model

Hao Cheng*, Hongji Shang*, Bin Lan, Houde Liu†, Xueqian Wang† and Bin Liang

*Abstract*— Continuum robots are typically slender and flexible with infinite freedoms in theory, which poses a challenge for their control and application. The shape sensing of continuum robots is vital to realise accuracy control. This letter proposed a novel general real-time shape sensing framework of continuum robots based on the piecewise polynomial curvature (PPC) kinematics model. We illustrate the coupling between orientation and position at any given location of the continuum robots. Further, the coupling relation could be bridged by the PPC kinematics. Therefore, we propose to estimate the shape of continuum robots through orientation estimation, using the off-the-shelf orientation sensors, e.g., IMUs, mounted on certain locations. The approach gives a valuable framework to the shape sensing of continuum robots, universality, accuracy and convenience. The accuracy of the general approach is verified in the experiments of multi-type physical prototypes.

## I. INTRODUCTION

Robots today are making a considerable impact on human life in many significant ways, from industrial manufacturing to medical application. As a novel concept, continuum robots have backbone structures with infinite degrees-of-freedom (DoF) theoretically, rather than joints used in traditional rigid robots [1]. Based on this concept, many continuum robot prototypes are produced: various backbone materials, e.g., polyamide material [9], elastic steel [10, 12], silicone (soft continuum robots) [11]; various actuators, e.g., pneumatic actuator [11], cable drive [10, 12]. The concept comes from bionics. Biologically inspired by snakes, the researchers hope to make slender and flexible robots, which can be used in narrow space from earthquake search and rescue (SAR) [5] to minimally invasive surgery (MIS) [6, 7]. However, the accurate modelling and control of continuum robots is still an open problem, as its high or even infinite freedoms.

Although extensive model-based research has been performed, accurate and useful shape estimation of continuum robots remains challenging, particularly in complex environment interactions [7]. The continuous deformable designs make continuum robots challenging to model their shape accurately. Meanwhile, real-time and accurate shape sensing is required to realise effective feedback control. Hence, a general and effective shape model and sensing approach are urgently required for various continuum robots prototypes.

* These authors contribute equally.   † Corresponding author.
This work was supported by National Natural Science Foundation of China (61803221 & U1813216), the Open Project of Shenzhen Institute of Artificial Intelligence and Robotics for Society (AC01202005004).
Hao Cheng, Hongji Shang, Bin Lan, Houde Liu (corresponding author) and Xueqian Wang (corresponding author) are with the Shenzhen International Graduate School, Tsinghua University, 518055 Shenzhen, China (E-mails: {Liu.hd, wang.xq}@sz.tsinghua.edu.cn).
Bin Liang is with the Department of Automation, Tsinghua University, 100084 Beijing, China.

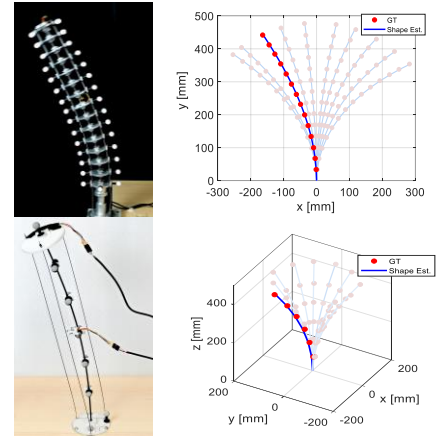

Figure 1. Multi-type physical prototypes of continuum robots, as well as the results of shape sensing by using proposed general framework approach. The top one is the planar case of spring steel backbones. The spatial case of the elastic rod is shown at the bottom.

The current shape sensing approaches of continuum robots are mainly based on sensors, such as cameras, optical fibres, electromagnetic (EM). Hannan et al. estimated the curvature of each segment from the external global image of the robot arm based on computer vision technology to obtain the overall manipulator shape [14]. In our previous work, a generalised epi-polar constraint for the end camera was derived; the parameter estimation can be realised by solving the equation [10]. S. Sareh et al. measured them by using optical fibres mounted on continuum robots [13]. A. Gao et al. proposed simultaneous shape and Tip force Estimation through optical fibre tension/shape/force sensing [15]. S. Huang et al. estimated the end position of cable-driven continuum robots by force sensors, establishing a pseudo-rigid body model for mechanical analysis [12]. For external sensors such as cameras, application scenarios are limited due to their dependence on external information. A case in point is that the global image cannot be got in unknown environments. In contrast, internal sensors are less sensitive to the environment and are suitable for a wide range of applications. Nevertheless, the current approaches by internal sensors usually require the design of complex sensors, e.g., fibre Bragg gratings (FBG) [16, 17], resistor shape sensors [18].

As developed internal sensors, inertial measurement units (IMU) are typically used in devices that require motion control, such as automobiles, unmanned aerial vehicles, and robots. Besides, the well-researched attitude solving approaches, such as error-state Kalman filter (ESKF) [19], provide accurate and real-time attitude estimation via the IMU. In hyper-redundant robots, Z. Zhang et al. estimated the shape of snake-like robots by mounting multiple IMUs on each segment [8]. However, the potential of the IMU for shape sensing of continuum robots needs further research; the challenge is how to bridge the gap between the pose estimation of certain points and the whole shape.

The lack of distinct links in continuum systems makes the standard robot manipulator modelling strategy of a finite number of coordinate frames (each fixed in one link) inappropriate for modelling [1]. To simplify the problem of robot state estimation and control, the piecewise constant curvature (PCC) model is widely used to describe the kinematics and dynamics of multi-segment continuum robots [28]. H. Wang et al. proposed an adaptive visual servo controller based on piecewise constant curvature (PCC) kinematic [20]. R. K. Katzschmann et al. proposed a control architecture under the hypothesis of PCC [11]. C. D. Santina et al. discussed the flaws of historical parametrisation under PCC assumption, and contributed an improved PCC state representation [21]. However, the PCC hypothesis is too idealistic to describe the actual continuum robot honestly. H. Cheng et al. proposed approximate piecewise constant curvature (APCC) equivalent model and deviation calibration approach to improving the PCC model [10]. C. D. Santina et al. put forward the kinematics model of the soft robot based on polynomial curvature, which accurately models the shape of continuum robots [22].

In this work, we further analyse the properties of parametrization on the shape of continuum robots, i.e. piecewise polynomial curvature (PPC), with a specific focus on its use in shape sensing through multi-attitude (orientation) solving. For rigid robots, the configuration is directly measurable, e.g. joint angles. Through the PPC kinematics model, we achieve the same in continuum robots by bridging the orientation and position, as well as estimating the shape via a limited number of orientation sensors, e.g. IMU, fibre optic gyroscope (FOG) [29]. We demonstrate the generality and effectiveness of our approach by using multi-type physical prototypes.

This letter contributes with
- A general shape sensing framework of continuum robots using orientation sensors (e.g., IMU, FOG), suited for multi-type continuum robots;
- An in-depth analysis of coupling between orientation and position of continuum robots via piecewise polynomial curvature (PPC) kinematics;
- Multi-type physical prototype experiments verifying the proposed approach by IMUs.

The rest of this paper is structured as follows. Sect. II provides the PPC kinematics model and illustrates the coupling relation between orientation and position of continuum robots. The general framework for shape sensing of continuum robots is presented in Sect. III. Multi-type physical prototypes are used to verify our approach in Sect. IV, along with analysis. Finally, this letter is concluded with a discussion and future work in Sect. V.

## II. BRIDGING THE ORIENTATION AND POSITION: PIECEWISE POLYNOMIAL CURVATURE KINEMATICS

In rigid robots, the map from configuration to pose can be accurately established through the D-H method [23]. Though the continuum robots have infinite DoF in theory, their control input is limited in practice. Therefore, the actual continuum robots are approximately subjected to specific ideal models in the structure design. For example, piecewise constant curvature (PCC) models have been proven as a very useful tool with a vast range of applications on design [18], sensing [25] and control [11, 21].

However, PCC models still have some severe flaws, i.e., the deviation between actual continuum robots and the over-simplification model, the influence of friction, gravity, and so on [10]. C. D. Santina et al. showed a polynomial curvature model in control applications, which considered the non-constant curvature issue [22]. Further, we generalise and obtain piecewise polynomial curvature (PPC) models, which are more accurate kinematics models. We will reveal that PPC models can bridge the orientation and position, similar to the D-H method, i.e., a coupling relation, which is the basis for shape sensing from orientation.

TABLE I.   LIST OF MAIN SYMBOLS

| Symbol | Property |
| --- | --- |
| $s_i \in [0,1]$ | Normalised coordinate along the $i$-th segment |
| $L_i \in \mathbb{R}^+$ | Backbone Length of the $i$-th segment |
| $\phi_i : \mathbb{R}^+ \to [0, 2\pi]$ | Bending direction of the $i$-th segment at a given time |
| $\alpha_i : [0,1] \times \mathbb{R}^+ \to [0, \pi]$ | Orientation of the i-th segment at a given location and a given time in bending plane |
| $q : [0,1] \times \mathbb{R}^+ \to \mathbb{R}$ | Local curvature at a given location and a given time [22] |
| $\boldsymbol{K}$ | Sequence space of curvature functions Mapping $\mathbb{N} \times \mathbb{R}^+$ into $\mathbb{R}$ [22] |
| $\Theta \in \boldsymbol{K}$ | Modal configuration at a given time |
| $m \in \mathbb{N}$ | Order of the approximation |
| $[\,\cdot\,]_m : \boldsymbol{K} \to \mathbb{R}^{m+1}$ | Truncation approximation operators |
| $\theta_k \doteq \Theta(k, \cdot) : \mathbb{R}^+ \to \mathbb{R}$ | $k$-th coordinate in modal space at a given time |
| $\theta \doteq [\Theta_i]_m : \mathbb{R}^+ \to \mathbb{R}^{m+1}$ | Reduced dimensionality description of $i$-th segment at a given time |

### A. PPC Kinematics

We introduce the piecewise polynomial curvature (PPC) models based on the seminal works: polynomial curvature [22]. In theory, the PPC kinematics provided an accurate model of the infinite-dimensions shape of 3D continuum robots. Through order truncation, approximating at any level of accuracy can be achieved. Fig. 2 shows a case of a PPC continuum robot.

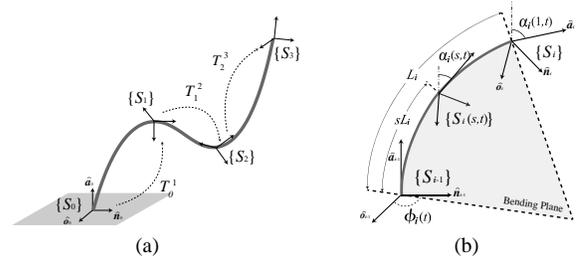

Figure 2.   The illustration of the PPC continuum robot. (a) An example of an overall PPC continuum robot with three PC segments. (b) shows the geometric relationship of an inextensible PC segment.

Similar to PCC models, the PPC kinematics consider the continuum robots composed by a sequence of a fixed number of segments are continuously deformable with polynomial curvature in space (PC) but variable in time, merged so that the whole space curve is everywhere differentiable. Instead of an over-simplification constant curvature curve, the PC curve is a more accurate depiction of the actual shape.

Focusing on basic problems, we consider the inextensible case in this work, which would be relaxed in future work. The main symbols used in this letter are defined as Table I. The PPC kinematics model is based on hypothesis as follows:

*Hypothesis 1:* The bending of arc in a certain segment occurs on the same plane, i.e. no twist.

*Hypothesis 2:* Continuous change of curvature, i.e. the curve is differentiable everywhere.

*Hypothesis 3:* The robot is inextensible, i.e. the length is constant.

We call $L_i \in \mathbb{R}^+$ the length of $i$-th segment. The curve is parameterised by a normalised coordinate $s \in [0,1]$, so that the point at coordinate $s$ is $sL_i$ far from the base of the segment. The curvature function $q_i(s,t)$ is introduced to specify the pose in the bending plane of the $i$-th segment. Through assuming this function to be analytic in $s$, the $q_i(s,t)$ can be expressed as an infinite polynomial (see [22] for more details):

$$q_i(s,t) = \sum_{k=0}^{\infty} \theta_k(t) s^k , \quad (1)$$

Where $\theta_k \in \mathbb{R}$ is the modal component of order $k$.

This function provides an exact description of the curve shape of $i$-th segment in the bending plane.

Based on the curvature function $q_i$, the local orientation in the bending plane of $i$-th segment is obtained by direct integration of the curvature

$$\alpha_i(s,t) = \int_0^s q_i(v,t)dv = \sum_{i=0}^{\infty} \theta_i(t) \frac{s^{i+1}}{i+1} , \quad (2)$$

Therefore, the Cartesian coordinates of each point along $i$-th segment in the bending plane as follow

$$\begin{cases} \dfrac{x_i(s,t)}{L_i} = \int_0^s \cos(\alpha(v,t))dv \\ \dfrac{y_i(s,t)}{L_i} = \int_0^s \sin(\alpha(v,t))dv \end{cases}, \quad (3)$$

We discuss its integrability in the next part. From (2), the contribution of the higher-order modal to orientation gradually decreases. Therefore, the finite-dimensional approximation of the polynomial curvature is reasonable in the form

$$q_i(s,t) \simeq \sum_{k=0}^{m} \theta_k(t) s^k , \quad (4)$$

When $m=0$, it will be the constant curvature model.

Combining orientation $\alpha_i(s,t)$ and bending direction $\phi_i(t)$, the homogeneous transformation of the $i$-th segment at a given location $s$ results

$$T_{i-1}^i(s,t) = R_a(\phi_i(t)) \begin{bmatrix} R_n(\alpha_i(s,t)) & \begin{bmatrix} x_i(s,t) \\ y_i(s,t) \\ 0 \end{bmatrix} \\ \begin{bmatrix} 0 & 0 & 0 \end{bmatrix} & 1 \end{bmatrix} R_a(-\phi_i(t)) . \quad (5)$$

Where $T_{i-1}^i(s,t)$ identify the three axes of $\{S_i(s,t)\}$ at a given location $s$ and a given time $t$, with coordinates expressed w.r.t. $\{S_{i-1}\}$.

*B. The Integrability of Cartesian Coordinates*

Eq. (3) is well known for not being integrable in closed form, which prevents an accurate and analytical dynamical description of the general case [22]. In this part, we prove the integrability of (3) and discuss the analytic solutions of zero-order and first-order approximations, i.e. $m = 0$ or 1.

*Lemma 1:* $\sin(\sum_{k=0}^{\infty} \theta_k \frac{s^{k+1}}{k+1}), \cos(\sum_{k=0}^{\infty} \theta_k \frac{s^{k+1}}{k+1}) \in R[0,1]$

*Proof:* $\sin(\sum_{k=0}^{\infty} \theta_k \frac{s^{k+1}}{k+1}), \cos(\sum_{k=0}^{\infty} \theta_k \frac{s^{k+1}}{k+1}) \in C[0,1]$ ∎

The thesis of Lemma 1 tells us that the integrals can be evaluated numerically at any level of precision. However, the numerical approach makes it hard to get further theoretical analysis. We briefly discuss analytic solutions to contribute ideas for future work, such as motion planning and control.

Under zero-order approximations, (3) changes to the standard trigonometric integral, equivalent to the constant curvature model. The result is

$$\frac{x_i(s,t)}{L_i} = \frac{1}{\theta_0} \sin(\theta_0 s), \quad \frac{y_i(s,t)}{L_i} = \frac{1}{\theta_0} (1 - \cos(\theta_0 s)) , \quad (6)$$

The first-order case involves trigonometric integrals with square terms, which is not integrable in closed form with the function of real variable. Fortunately, it has a closed-form with complex function, using Fresnel integral that has been well researched in areas such as optics:

$$\begin{cases} f_S(x) = \int_0^x \sin \frac{\pi}{2} v^2 dv = \frac{\sqrt{\pi}}{4} (\sqrt{i}\,\mathrm{erf}(\sqrt{i}\,x) + \sqrt{-i}\,\mathrm{erf}(\sqrt{-i}\,x)) \\ f_C(x) = \int_0^x \cos \frac{\pi}{2} v^2 dv = \frac{\sqrt{\pi}}{4} (\sqrt{-i}\,\mathrm{erf}(\sqrt{i}\,x) + \sqrt{i}\,\mathrm{erf}(\sqrt{-i}\,x)) \end{cases}, \quad (7)$$

Where $\mathrm{erf}(\cdot)$ is the Gauss error function. The result of the integration is

$$\begin{cases} \dfrac{x_i(s,t)}{L_i} = \int_0^s \cos(\theta_0 v + \frac{1}{2}\theta_1 v^2)dv \\ \quad = \sqrt{\frac{\pi}{\theta_1}} \cdot (d \cdot (f_C(a) - f_C(b)) + c \cdot (f_S(a) - f_S(b))) \\ \dfrac{y_i(s,t)}{L_i} = \int_0^s \sin(\theta_0 v + \frac{1}{2}\theta_1 v^2)dv \\ \quad = \sqrt{\frac{\pi}{\theta_1}} \cdot (d \cdot (f_S(a) - f_S(b)) - c \cdot (f_C(a) - f_C(b))) \end{cases}. \quad (8)$$

Where $a(s) = \dfrac{\theta_0 + \theta_1 s}{\sqrt{\pi \theta_1}}$, $b = \dfrac{\theta_0}{\sqrt{\pi \theta_1}}$, $c = \sin \dfrac{\theta_0^2}{2\theta_1}$, $d = \cos \dfrac{\theta_0^2}{2\theta_1}$.

*C. On the Coupling between Orientation and Position*

In rigid robots, the relationship between orientation and position can be established by the D-H method. However, with infinite DoF in theory, it is difficult to establish such a coupling relationship accurately for continuum robots.

Under the PPC kinematics, the modal configuration $\Theta \in K$ builds the bridge between the orientation and position. We consider the finite-dimensional approximation of curvature function (4) in $m$ order

$$\Theta = [\theta_0, \ldots, \theta_m]^T , \quad (9)$$

Then select $m$ different locations of a certain segment

$$s = [s_0, \ldots, s_m]^T , \quad (10)$$

Where $0 < s_0 < s_1 < \ldots < s_m \leq 1$.

We assume that the orientations at these positions are known at a given time as $\alpha_i(s_k,t)$, $k \in \{0,...,m\}$. From (2), the $m$-order modals and orientations at $m$ locations satisfy the linear equation $\mathbf{A}\Theta = \mathbf{b}$ as follow

$$\begin{bmatrix} s_0 & \frac{1}{2}s_0^2 & \cdots & \frac{1}{m+1}s_0^{m+1} \\ \vdots & \vdots & & \vdots \\ s_m & \frac{1}{2}s_m^2 & \cdots & \frac{1}{m+1}s_m^{m+1} \end{bmatrix} \begin{bmatrix} \theta_0 \\ \vdots \\ \theta_m \end{bmatrix} = \begin{bmatrix} \alpha(s_0,t) \\ \vdots \\ \alpha(s_m,t) \end{bmatrix}. \quad (11)$$

Thus the modal configuration can be obtained from the orientations by solving the equation. Furthermore, we show that (11) has a unique solution.

*Lemma 2:* $\mathbf{A}\Theta = \mathbf{b}$

*Proof:* The determinant of the coefficient matrix $\mathbf{A}$

$$\det(\mathbf{A}) = \det\left( \begin{bmatrix} s_0 & \cdots & s_0^{m+1} \\ \vdots & & \vdots \\ s_m & \cdots & s_m^{m+1} \end{bmatrix} \begin{bmatrix} 1 & & \\ & \ddots & \\ & & \frac{1}{m+1} \end{bmatrix} \right)$$

$$= \frac{\prod_{k=0}^{m} s_k}{(m+1)!} \cdot \prod_{0 \le j < i \le m}(s_i - s_j) > 0$$

The matrix $\mathbf{A}$ is invertible; the linear equations have a unique solution $\Theta = \mathbf{A}^{-1}\mathbf{b}$. ∎

Further, the pose at any location $s \in [0,1]$, including orientation and position, could be obtained by solving (2) and (3), respectively.

Thus, the functional relationship between the orientation at $m$ locations and the pose at any location is established, mapping some orientations to the modal configuration and finally the whole pose. Next section, we will discuss its application in the shape sensing of continuum robots.

### III. A GENERAL FRAMEWORK FOR SHAPE SENSING OF CONTINUUM ROBOTS

This section introduces the shape sensing of continuum robots and gives its general real-time approach framework by using orientation sensors such as IMU and FOG.

#### A. Problem Description and Approach

Real-time accurate shape sensing is a significant basis to realise the accurate feedback control and obstacle avoidance planning of continuum robots. The proposed approach is based on the orientation sensors mounted on the continuum robot under the PPC kinematics in $m$-order approximation.

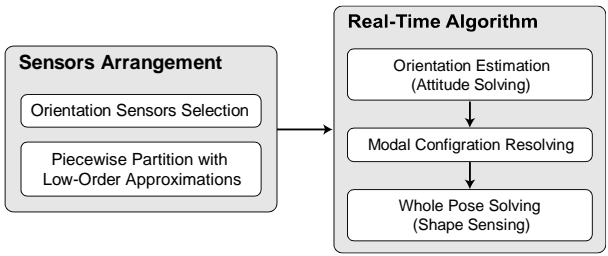

Figure 3. A diagram illustrating the pipeline of proposed approach.

The pipeline of the proposed shape sensing framework is shown in Fig. 3. The approach starts with hardware preparation, i.e., determining the sensors arrangement, which contains selecting orientation sensors and the piecewise partition with low-order finite-dimensional curvature approximations. The following is the real-time algorithm, including orientation estimation (attitude solving), modal configuration resolving and the whole pose solving (shape sensing).

In sensors arrangement, it needs to be considered with the mechanical structure of actual continuum robots. For sensors selection, IMU sensors offer proven solutions. Besides, orientation sensors can also be designed according to the actual situation, e.g., soft robots with fibre embedding as FOG sensors [29]. For partition and curvature approximations, it is necessary to consider the structure of the actual robot and the motion situation, which would be discussed in IV.A with specific robot prototypes. Details of the real-time shape sensing algorithm are shown as following.

#### B. The Real-Time Shape Sensing Algorithm

For orientation estimation, there are mature algorithms, such as ESKF [19], Mahony [26] and Madgwick [27].

In orientation algorithms, quaternions are usually used to express the orientation for efficient calculation [19, 26, 27]. We now illustrate that the quaternion representation is more intuitive for the pose of continuum robots under the PPC kinematics.

Under hypothesis 1, the orientation at a given location of a certain segment is only 2-DoF, which can be succinctly expressed in the form of axis-angle, i.e. vertical axis $\mathbf{n}$ of the bending plane and orientation $\alpha(s,t)$ in the bending plane

$$\alpha(s,t), \quad \mathbf{n} = \begin{bmatrix} n_x \\ n_y \\ n_z \end{bmatrix} = \begin{bmatrix} -\sin\phi(t) \\ \cos\phi(t) \\ 0 \end{bmatrix}, \quad (12)$$

From the definition of quaternion

$$\mathbf{q} = [w, x, y, z]^T = [\cos(\tfrac{\alpha}{2}), n_x\sin(\tfrac{\alpha}{2}), n_y\sin(\tfrac{\alpha}{2}), n_z\sin(\tfrac{\alpha}{2})]^T, \quad (13)$$

The configuration can be expressed as

$$\begin{cases} \alpha(s,t) = 2\arccos(w) \\ \phi(t) = -\arctan(\tfrac{x}{y}) \end{cases}. \quad (14)$$

Note that $z$ is zero for no twist, which holds true for actual cases, as shown in IV. Under the $m$-order approximation, the orientation at $m$ locations is obtained by sensors in location $s$, i.e. $\mathbf{q}(s_k,t)$, $k \in \{0,...,m\}$. From (11), (14), the modal components can be resolved by solving linear equations.

With the real-time modal configuration, the whole pose of continuum robots, i.e. the orientation and position at any location, can be obtained by using the $m$-order approximate PPC kinematics. Eq. (4) provides orientations. Then the position can be obtained by the integration (2). Note that due to the complexity of solving the integral, the position can only be obtained by numerical integration (also real-time) for the higher-order curvature approximations, i.e. $m \ge 2$; the 0-order and 1-order approximations can be calculated by the analytical equations (6) and (8), respectively. We will illustrate that the first-order approximation is sufficient for most cases, with a reasonable sensors arrangement.

The above algorithm can estimate the shape in real-time. Specifically, orientation sensors, such as the IMU, have fast sampling rates, as well as orientation algorithms with the real-time performance [19, 26, 27]. In modal and pose solving, linear equations and integrals can also be solved in real-time. Note that parallel computing would improve the computational efficiency by such multiple segments.

## C. Error Analysis

There are two primary sources of error in shape estimation. The first one is the error propagation of orientation estimation; the other is the curvature modal approximation error. We briefly discuss the influence of both error sources on the accuracy of shape sensing, which would be further researched in the next work.

For the error propagation, we analyse the error of position in the bending plane propagating by orientation estimation, which is based on the quaternion extended Kalman filtering (EKF). For the sake of space and readability, we consider the case of first-order curvature approximation, i.e. $m = 1$, which needs two sensors in the location $s_0$ and $s_1$, respectively. We assume $w_0$, $w_1$ are quaternion constant terms of $s_0$ and $s_1$.

According to (3), (11) and (14), the uncertainty equation of the position is

$$\sum\nolimits_{postion}(s,t) = J_w^p \Sigma_w(t) (J_w^p)^T, \quad (15)$$

Where the Jacobian matrix $J_w^p = J_m^p J_w^m$ represents the first-order differential of the mapping from the constant term of quaternion $w$ to the Cartesian coordinates in the bending plane. $\Sigma_w(t) = \text{diag}(\sigma_{w0}, \sigma_{w1})$ is combined covariance matrix, combining with the variances of $w_0$ and $w_1$.

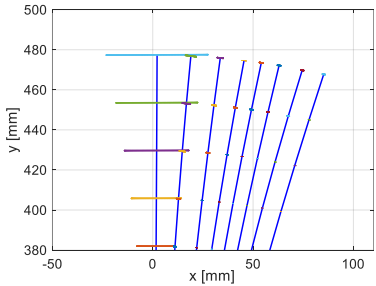

Figure 4. An illustration of uncertainty ellipse. The blue curves are the shape of robots, and the rest are the distribution of Uncertainty Ellipse.

For the error generated by curvature modal approximation, we will discuss it in IV.B by showing examples where the first-order approximation cannot represent the curve due to the influence of external forces.

## IV. MULTI-TYPE PHYSICAL PROTOTYPES EXPERIMENTS AND ANALYSIS

In this section, we validate the shape sensing approach presented in the previous sections through physical experiments on two types of actual prototypes. We start by first discussing the strategy for sensors configuration with multi-type prototypes. This is followed by the results of experiments and related analysis. In all experiments, the ground truth of shapes is represented by markers, which collected by the motion capture system (NaturalPoint, Inc. DBA OptiTrack) with a specified accuracy of 0.20-mm RMS.

## A. Strategy for Sensors Arrangement

For the specific prototypes of continuum robots, sensors selection is required firstly for considering the factors of hardware and electrical interface. As mature devices, the IMU provides a highly integrated orientation sensor solution, the preferred sensors. Nevertheless, for the cases that are difficult to fix IMUs, e.g., soft continuum robots, a better solution is the FOG made by embedding fibre optics [29]. The selection of piecewise partition and the modal order of curvature approximation is equally significant to achieve better shape sensing. In general, the piecewise partition mainly depends on the partition of controllable segments, e.g. the cable-driven segment, which would be further studied in future work.

We select the approximate modal order from low to high according to the motion of each segment. In the subsequent experiments, we illustrate that the first-order approximation is generally applicable in most cases, including the scenario with tip-environment interaction. Nevertheless, as for external force in the middle of the segment, the higher-order approximation is necessary.

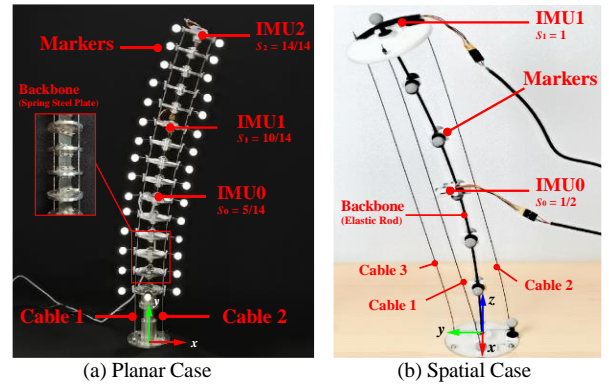

(a) Planar Case      (b) Spatial Case

Figure 5. The two types of cable-driven prototypes for experimental validation. (a) the planar prototype consists of 14 sub-segments connected by spring steel plates with two cables. (b) the spatial prototype consists of a whole elastic rod with three cables.

Three and two IMUs (ICM-20948) were mounted on the planar and spatial prototype, respectively. This IMU is a 9-axis mini-type sensor, recording at 60Hz in experiments.

In the following, we verify the validity of the shape sensing by using elastic backbones. Meanwhile, we will discuss the effect of external forces on curves. Specifically, we compare the accuracy of the 1-order (using IMU0 & IMU2) and 2-order (using all IMUs) approach in various conditions. In spatial case, we focus on the accuracy of bending direction.

## B. Planar Case: Spring Steel Plate Backbone

In the planar case, we mainly verify the description ability of shapes in the bending plane. The experimental setup consists of a planar spring steel prototype of continuum robots and the motion capture system. This prototype's structure is illustrated at the left of Fig.5, which consists of 14 sub-segments by a series of disks connected by spring steel plates. The length of this prototype is 480mm. We mounted three IMUs on the tip and body locations, i.e. $s_0 = 5/14$, $s_1 = 10/14$, $s_2 = 1$, used to 1-order and 2-order approach. Specifically, the results of 1-order come from the IMUs at $s_0$ and $s_1$, while the 2-order results from all.

TABLE II. ESTIMATION ERRORS OF THE MULTI-TYPE PROTOTYPES

| Test | | Planar Case | | | | | Spatial Case |
|---|---|---|---|---|---|---|---|
| | | Swing | Free Oscillation | Interaction (Tip) | Interaction (Body) | | Circular Motion |
| | | | | | 1-Order Estimation | 2-Order Estimation | |
| Shape RMSE [a] | Mean (mm) | 2.64 | 2.94 | 2.69 | 4.30 | 2.47 | 4.35 |
| | SD (mm) | 0.60 | 0.97 | 0.52 | 1.79 | 0.68 | 1.31 |
| | Max (mm) | 5.78 | 7.41 | 4.51 | 12.13 | 5.06 | 8.63 |
| | BRE (%) [d] | 1.20 | 1.54 | 0.94 | 2.53 | **1.05** [e] | 1.77 |
| Tip Abs. Error [b] | Mean (mm) | 2.52 | 3.75 | 2.97 | 6.20 | 1.78 | 4.78 |
| | SD (mm) | 0.93 | 2.23 | 1.39 | 4.33 | 1.06 | 2.10 |
| | Max (mm) | 6.76 | 12.28 | 8.80 | 24.83 | 6.41 | 10.03 |
| | BRE (%) [d] | 1.41 | 2.56 | 1.83 | 5.17 | **1.34** [e] | 2.05 |
| Bending Direction Abs. Error [c] | Mean (deg) | - | - | - | - | - | 1.09 |
| | SD (deg) | - | - | - | - | - | 1.02 |
| | Max (deg) | - | - | - | - | - | 5.07 |

[a] It means the root mean square errors (RMSE) of estimation and experiment at multiple even-distributed locations. [b] It means the absolute error between the estimation and the measured tip position. [c] It means the absolute error between the estimation and the measured bending direction, which only considered in the spatial case. [d] The percentage is the difference between the max error to the robots' length, which both are 480mm. [e] The accuracy of the 2-order is higher than that of the 1-order approach in complex interaction.

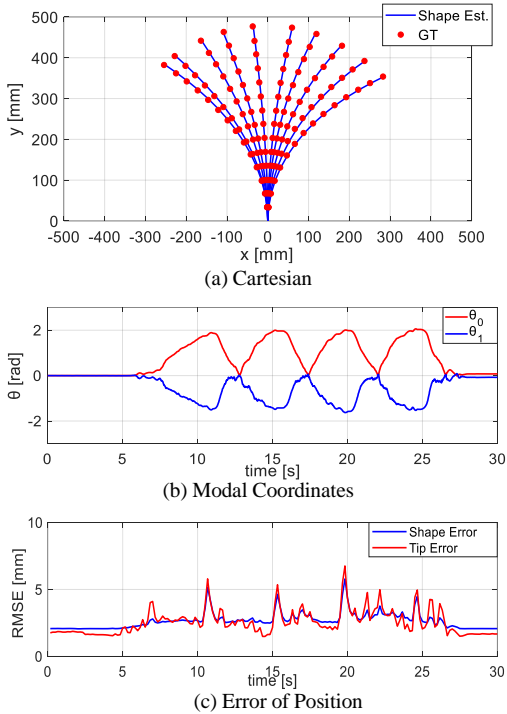

Figure 6. The experiment of swing. (a) shows the result in Cartesian coordinates. (b) gives the modal coordinates estimated by our approach vs. time. (c) shows the RMSE of the whole shape and tip.

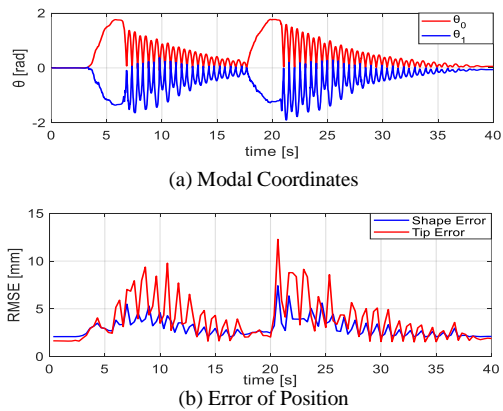

Figure 7. The experiment of free oscillation, demonstrating the dynamic performance of our approach.

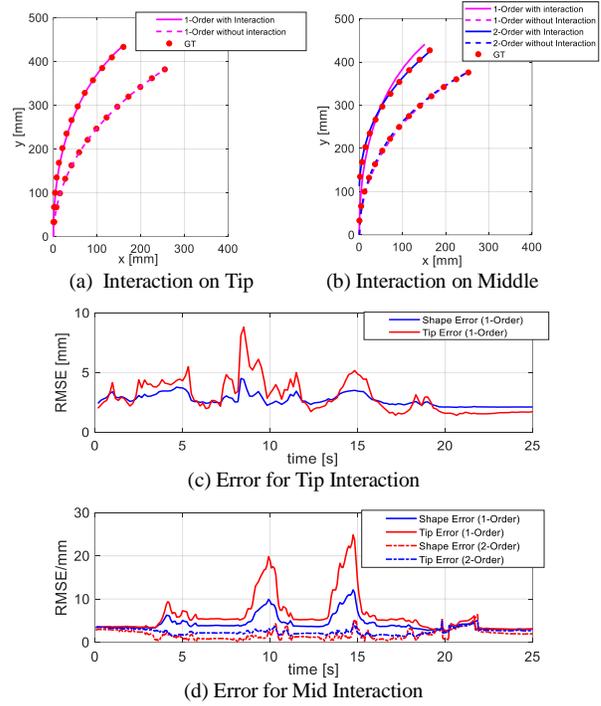

Figure 8. Shape estimation under external forces interaction. The external forces applied at the tip and middle. (a) and (b) show the effect on the curve shape for external forces on tip and middle, respectively. (c) illustrates that the first-order approximation is still valid with the tip forces. (d) shows that the higher-order is necessary for mid interaction.

Three motion tests were selected to verify our approach, i.e. Swing, Free Oscillation and Interaction with External Forces: 1) the prototype moves slowly back and forth; 2) the prototype moves to the maximum then release and swings freely, which demonstrates the dynamic performance of shape sensing; 3) the prototype moves to the maximum then interacts with external forces at tip and body. Results are shown in Fig. 6, Fig.7 and Fig.8 for each test. Detailed error measurements are listed in Table II.

There are two terms to quantify the error, including the mean error of tip tracking and the mean error of robot shape [8]. Specifically, we calculate the root mean square errors (RMSE) of estimation and experiment at multiple even-distributed locations. Besides, the tip position error is described by the absolute error. Furthermore, the accuracy of the shape estimation approach is analysed by mean, variance, maximum value and bounds of relative error (BRE) in each motion. The BRE is relative to its length, i.e. 480 mm.

In the interaction case, we discuss the validity of this approach under interaction with external forces as Fig. 8. For tip interaction (common in applications), it has little effect on shape sensing accuracy. However, high-order shape sensing is needed for body interaction.

### C. Spatial Case: Elastic Rod Backbone

In the space case of elastic backbones, the experimental setup consists of a whole elastic rod prototype of continuum robots. The experimental prototype is shown at the right of Fig.5. The length of this prototype is 480mm. We mounted two IMUs on the location of tip and middle, i.e. $s_0 = 0.5$, $s_1 = 1$. Results are shown in Fig. 9. Detailed error measurements are listed in Table II.

In this case, we focus on the perception of the bending direction. Note that the bending direction has a singularity in origin, i.e. no bending, so that it is unstable near this singularity. When discussing the error of bending direction, we only consider the period during the deflection direction is clear, i.e. from 6.5 s to 49.5 s.

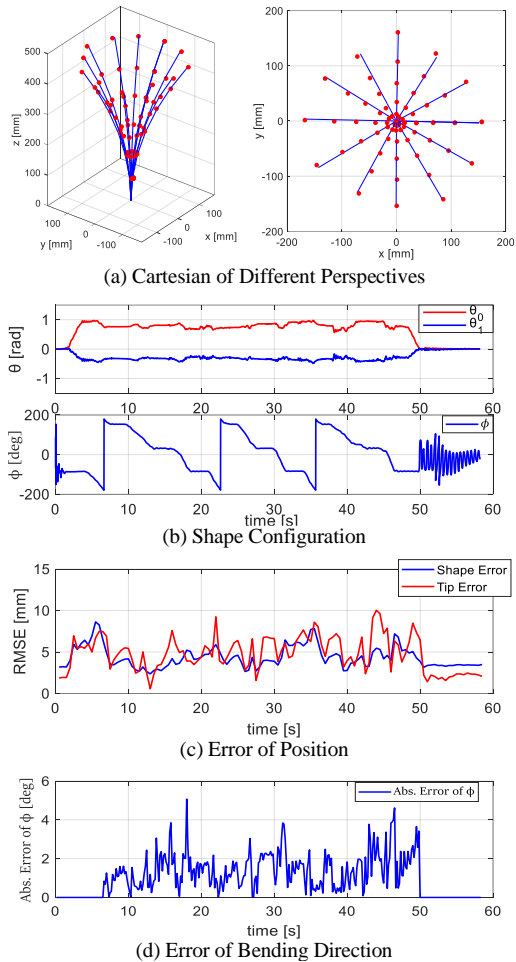

Figure 9. An elastic-rod continuum robot draws circles in space.

## V. CONCLUSION AND FUTURE WORK

This letter introduces a general shape sensing framework of continuum robots by using orientation sensors, such as inertial measurement units (IMU) and fibre optic gyroscope (FOG). We utilise the framework to attack the real-time shape sensing of multi-type continuum robots, which provide perceptual information for further application, such as feedback control and obstacle avoidance planning.

Future work will be devoted to applying the shape sensing approach to the feedback control of actual continuum robots. We will also test our approach in multi-segment continuum robots, and soft continuum robots with fibre embedding as FOG sensors. Besides, further error analysis will be studied.